\documentclass{article}
\usepackage{amssymb, graphicx, tabularx}
\usepackage{booktabs}

\usepackage[final]{corl_2025} %

\newcommand{\method}{ReasonNav}

\title{Human-like Navigation in a World Built for Humans}

\author{
  Bhargav Chandaka$^*$
  \And
  Gloria X. Wang$^*$
  \And
  Haozhe Chen
  \And
  Henry Che
  \And
  Albert J. Zhai
  \AND
  Shenlong Wang\\\\
  University of Illinois Urbana-Champaign\\
}

\begin{document}
\maketitle

\def\thefootnote{*}\footnotetext{Equal contributions.}\def\thefootnote{\arabic{footnote}}

\begin{abstract}
    When navigating in a man-made environment they haven’t visited before—like an office building—humans employ behaviors such as reading signs and asking others for directions. 
    These behaviors help humans reach their destinations efficiently by reducing the need to search through large areas. Existing robot navigation systems lack the ability to execute such behaviors and are thus highly inefficient at navigating within large environments. We present \method{}, a modular navigation system which integrates these human-like navigation skills by leveraging the reasoning capabilities of a vision-language model (VLM). We design compact input and output abstractions based on navigation landmarks, allowing the VLM to focus on language understanding and reasoning. We evaluate \method{} on real and simulated navigation tasks and show that the agent successfully employs higher-order reasoning to navigate efficiently in large, complex buildings. Project website: \href{https://reasonnav.github.io/}{https://reasonnav.github.io/}.

\end{abstract}


\section{Introduction}
\label{sec:intro}

Imagine that you are an office worker and are asked to deliver a report to Jane Doe's office. What steps would you take to complete this task? First, you might search in a directory to find out the building and room number for Jane Doe's office. Then, you might look for signs that indicate the direction of that room. You can integrate the information you receive from each sign with the layout of the scene you see around you to decide where to look next. Along the way, you might ask people nearby for further clarifications.
    
Our civilization is built to be easy for humans to navigate within. There is an abundance of knowledge-offering resources around us that we leverage to navigate the world efficiently. Directional signs are placed deliberately at junctions to eliminate the risk of going the wrong way. Room labels follow orderly patterns so that reading a few can allow one to infer the locations of other rooms. Such guidance is necessary in order to deal with the inherent uncertainty of navigation in unseen environments. 
    
Existing robot navigation systems lack the skills needed to leverage these resources and thus lose out in navigation efficiency by spending unnecessary time exploring. We call these skills, which include sign reading and asking for directions, \textit{higher-order navigation skills} because they require higher-order reasoning abilities and language processing. These skills become increasingly important in larger environments, where exploring in the wrong direction can cost a massive amount of time. 

Our key insight is that such higher-order navigation skills can be integrated in a unified manner by taking advantage of recent advances in large vision-language models (VLMs). In this paper, we present \method{}, a modular system for human-like navigation that leverages the zero-shot reasoning capabilities of a VLM in an agentic manner. The system is comprised of two streams: a low-level stream that handles localization, mapping, and path planning, and a high-level stream where the VLM performs high-level planning on abstracted observation and action spaces. Specifically, we represent the environment using a memory bank of landmarks (e.g. map frontiers, doors, people, signs) with attached textual information. This simplifies both the input and output spaces for the VLM agent, allowing it to focus on higher-order reasoning.

We evaluate \method{} in real and simulated environments. In both cases, the robot is tasked with finding a given room in a large (unseen) building. This mimics a practical indoor delivery scenario. We show that our abstraction design allows the VLM to interpret information from signs and people and use it to guide its decision-making. We compare our full system with ablated versions and demonstrate that such higher-order navigation skills greatly impact navigation performance. Overall, the results suggest that our VLM agent framework is a promising path forward for achieving human-like navigation efficiency using higher-order reasoning skills.

\begin{figure}
    \centering
    \includegraphics[width=1\linewidth,trim={48px 15px 48px 80px},clip]{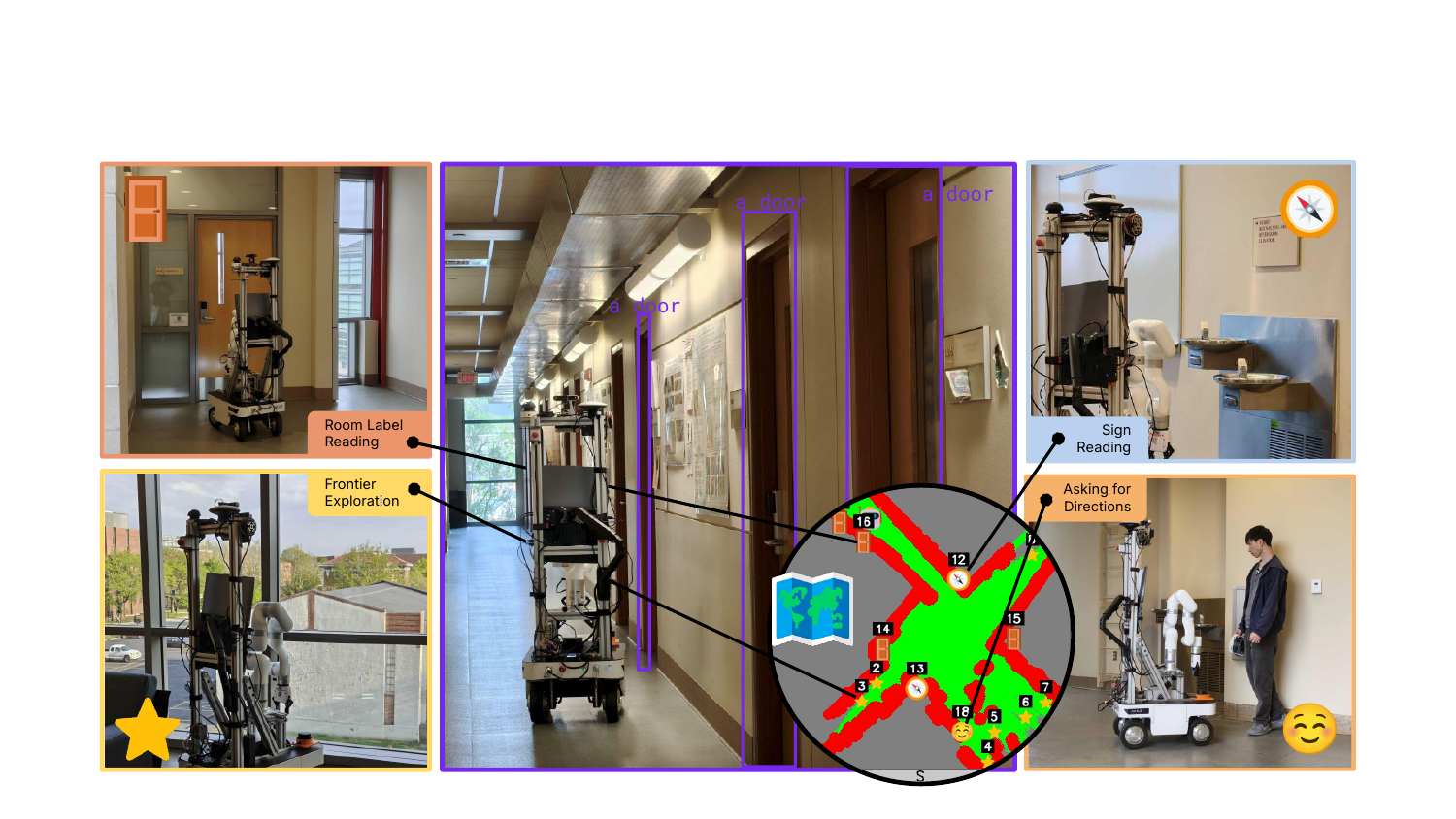}
    \vspace{-12pt}
    \caption{\textbf{Higher-order navigation skills.} Humans employ various skills involving higher-order reasoning in order to navigate to their destinations efficiently. These skills take advantage of key knowledge resources in the surrounding environment through high-level language and visual processing. We present a navigation method that imbues robots with these skills by integrating them in a VLM agent framework.}
    \label{fig:navskill}
    \vspace{-8pt}
\end{figure}

\section{Related Work}
\label{sec:related}

\paragraph{Agentic Foundation Models in Robotics.}
Task and motion planning (TAMP) approaches traditionally rely on predefined symbolic reasoning or optimization to plan for long-horizon tasks. Previous works \cite{huanginnermonologue, shahlm-nav, renrobotsthat, rajvanshi2024saynav} have leveraged large language models (LLMs) to decompose high-level instructions into actionable subtasks, allowing for more user-friendly robotics systems. More recent approaches utilize Vision-Language Models (VLMs) to ground reasoning for more general and capable robot systems. VLMs have been shown to generalize across diverse objects and tasks in table-top manipulation \cite{jiangroboexp, zawalski2024roboticcontrol, huang2023voxposer}, and enable zero-shot navigation to semantic goals across different environments \cite{longinstructnav, chang2024goat, shah2023gnm}. Integrating these capabilities for mobile manipulation has seen improved potential in recent works \cite{shah2024bumble, zawalski2024roboticcontrol, chiang2024mobilityvla}, which are divided into two main categories: 1. prompt-based querying and 2. fine-tuning for direct perception to action pipeline.
Our approach falls into the first category, querying a VLM for high-level task planning and using modular out-of-the-box controllers to execute actions. However, in contrast to the aforementioned methods that mainly rely on sensory inputs to perceive the world, our methods can leverage other resources, such as asking humans for help or actively seeking visual cues for navigation.

\paragraph{Open-world Navigation. }
In recent years, the rise in popularity of large-scale pre-trained foundation models has witnessed the emergence of open-world navigation. Particularly, early works of CLIP-on-Wheels (CoW) \cite{gadre2022cow} and LM-Nav utilize CLIP \cite{clip} to establish a top-down confidence map for language-guided object-goal navigation or to pre-compute a language-embedded topological graph. Further works have expanded on this direction, using foundation models to pre-build highly expressive language-embedded semantic maps for long-horizon and fine-grained navigation tasks \cite{werby23hovsg, huang23vlmaps, chen2023open, gu2024conceptgraphs, _Mahi_Shafiullah_2023, Peng_2023_CVPR, conceptfusion}. However, these approaches are computationally expensive, typically requiring multiple traversals over the operational area and hours of computing, and are unable to operate in unknown environments. Recent works \cite{longinstructnav, rajvanshi2024saynav, chen2023a2navactionawarezeroshotrobot, wu2024voronavvoronoibasedzeroshotobject, cai2023bridgingzeroshotobjectnavigation, Yu_2023, zhou2023esc} address this shortcoming by adopting LLMs and VLMs' high-level planners, taking advantage of their high-level reasoning capabilities to relax the requirement of costly pre-built maps. Our method falls into this category, enjoying the scalability and zero-shot transferability to the unknown world. However, we focus on practical navigation in man-made environments and the unique skills needed to succeed in such settings.

\paragraph{Interactive Navigation.}
Although these large foundation models have been trained on the vast majority of internet data and have shown promising results for robotic tasks, solely relying on them has proven to be inefficient. In recent years, the robotics community has been exploring human-in-the-loop feedback for corrections during robot's execution, especially for manipulation tasks \cite{zha2024distilling, onlinelangcorrection, li2023itp, liu2023interactive} and visual question answering tasks \cite{ren2024exploreconfident, renrobotsthat, thomason:corl19, dai2023think}. 
Despite showing promising results, these methods typically require immediate human feedback, which is often not possible in real-world navigation scenarios. In contrast, our work mitigates this issue by leveraging more than just human feedback as an additional source of information, utilizing wayfinding cues (room labels, navigation signs, web searches) for more robust and efficient navigation.

\section{Method}
\label{sec:method}

\begin{figure}
    \centering
    \includegraphics[width=\linewidth,trim={48px 55px 48px 45px, clip}]{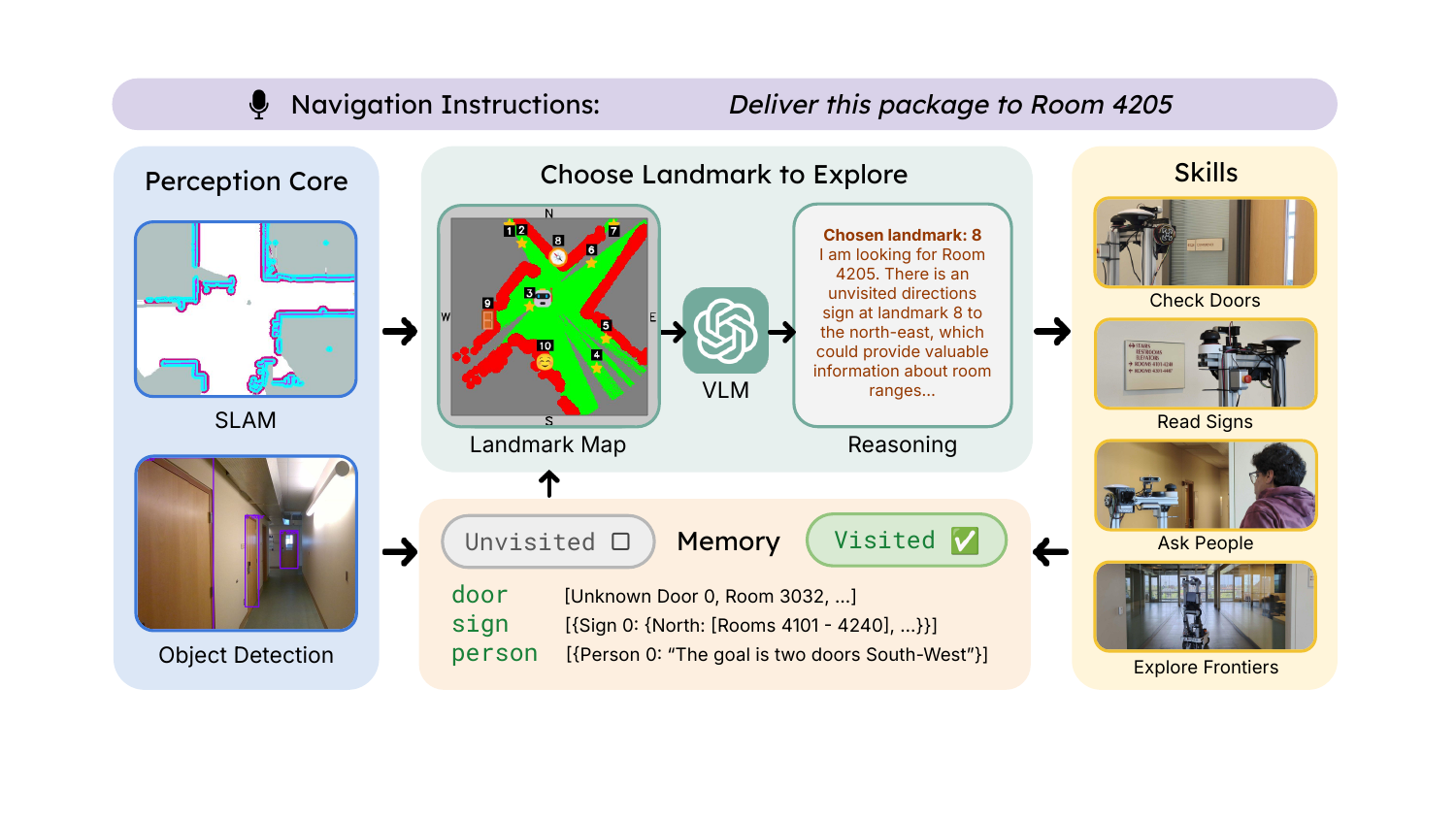} 
    \vspace{-12pt}
    \caption{\textbf{Overview of \method{}.} The system is comprised of a low-level stream and a high-level stream. The low-level stream performs SLAM and object detection for key object categories (doors, signs, and people), feeding into a global memory bank. The high-level stream consists of a VLM planner that receives abstracted observations in the form of a JSON landmark dictionary and a map visualization. The VLM outputs the next landmark to explore, upon which predefined behavior primitives are executed based on the landmark category. 
    }
    \label{fig:overview}
\end{figure}

\method{} is a modular system for human-like navigation that heavily leverages the zero-shot reasoning capabilities of a Vision-Language Model (VLM) for efficient exploration in unseen buildings. The system can be separated into a low-level stream and a high-level stream. The low-level stream includes standard localization and mapping modules that run at high frequency and an analytical path planner (Appendix~\ref{sec:lowlevel}). The high-level stream consists of a VLM agent that receives specially abstracted scene information to mimic the conscious decision-making processes used by humans during navigation (Sec.~\ref{sec:abstraction}). The VLM chooses map frontiers to explore and decides when to perform skills such as sign reading, which are executed via predefined behavior primitives (Sec.~\ref{sec:primitives}). 

\subsection{VLM Observation and Action Abstraction}
\label{sec:abstraction}
The key idea of our approach is to leverage VLMs in an agentic framework to integrate human-like behaviors that greatly improve navigation efficiency. VLMs excel at understanding language and conducting many forms of commonsense reasoning. However, they struggle at understanding complex spatial data and directly producing precise numerical outputs~\cite{fu2024blink, lumathvista}. Thus, we need to carefully design abstractions for both the input (observations) and output (actions) of the VLM in order to effectively leverage its reasoning capabilities. 

\paragraph{Landmarks.} Our abstraction design is centered heavily on the concept of \textit{landmarks}, which refer to salient objects that are especially important in navigation tasks. Specifically, the landmarks refer to objects of the three categories mentioned above: doors, people, and directional signs, along with frontiers of the top-down map. Our system populates a memory bank of the objects from the output of a detector and attaches additional navigation-relevant information to each one as various skills are performed (see Fig.~\ref{fig:overview}). For doors, we attach the text of the associated room label. For people, we attach a summary (generated by the VLM) of the information received from them. For directional signs, we attach a list of cardinal directions and the sign text reading(s) associated with each direction. All of the objects are attached to a label of ``Visited'' or ``Unvisited''.    

\paragraph{VLM Input and Output.} We prompt the VLM with text instructions and two forms of abstracted scene information. One is the memory bank of landmarks, including both objects and map frontiers, in JSON format. Each landmark is assigned an index and may have additional information attached as described above. The second form of information is an image visualization of the agent's current top-down map. The map is colored based on occupancy and explored areas, and for each landmark, we plot its location on the map with a symbol of its category and its index number (see Fig.~\ref{fig:navskill}). This gives a compact, high-level summary of the scene layout and the important objects the agent has seen thus far. We prompt the VLM to use these two forms of information to decide \textit{which landmark} to visit next. This design ensures that the VLM can flexibly choose any reasonable high-level plan while not being tasked with predicting precise numerical coordinates. The full prompt and detailed examples of the reasoning process can be found in Appendix ~\ref{sup:prompts}.

\subsection{Behavior Primitives}
\label{sec:primitives}
Each landmark category has an associated behavior primitive, which will be executed based on the VLM's choice. We describe each one below:

\paragraph{Frontier (Exploration).} The agent moves to the desired frontier and turns 360-degrees to scan its surroundings with Nav2's point-goal planner and controller~\cite{macenski2020marathon2}. Frontier navigation enables us to explore unvisited regions and identify more landmarks. 

\paragraph{Door (Room Label Reading).} The agent approaches the door and pans its camera while querying the object detector for \texttt{room label}. If a room label is detected, the agent moves closer and reads it via another call to the VLM. The text is attached to the door in the memory bank. If the goal is found, the episode ends here.

\begin{figure}
    \centering
    \includegraphics[width=.9\linewidth]{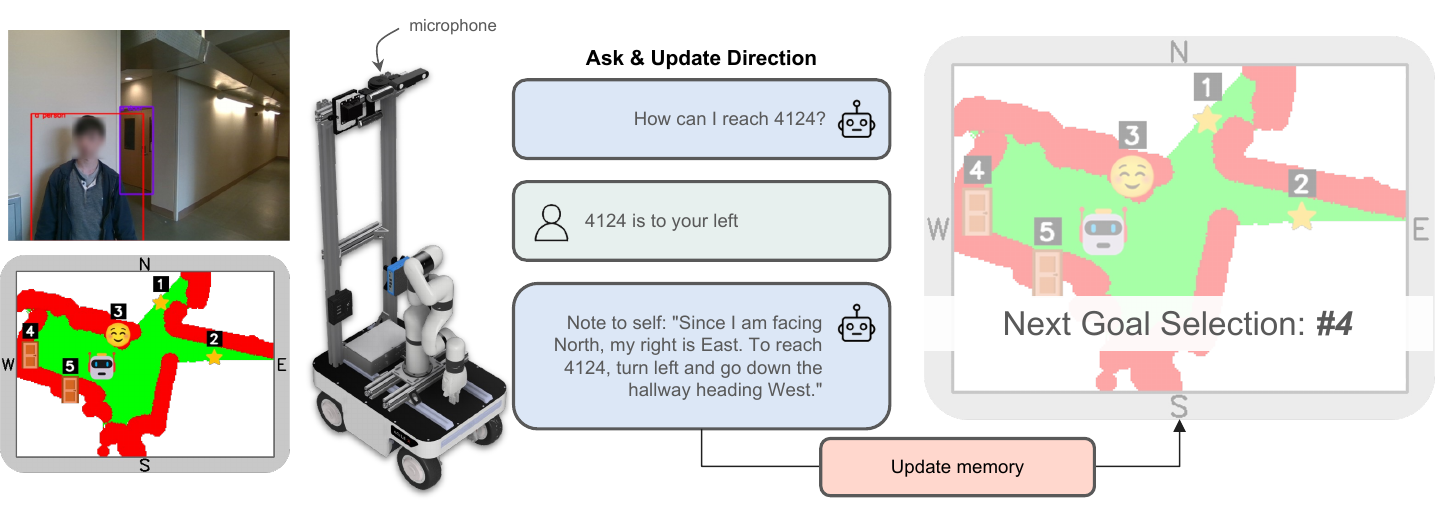}
    \caption{{\bf Overview of the “Direction Asking” Skill:} The agent identifies nearby humans and logs them in its spatial memory (\#3 in the map). When needed, it approaches and asks for goal directions via text-to-speech. The human’s verbal response is transcribed and updated in memory, enabling a more informed search towards the target (\#4) that avoids unvisited areas (frontiers in \#1 and \#2) unrelated to the goal and improves efficiency.}
    \label{fig:convo}
    \vspace{-.5em}
\end{figure}

\paragraph{Person (Asking for Directions).} The agent approaches the person and asks for directions using a text-to-speech model. It then records the person's response using speech-to-text. Next, it calls the VLM to produce a short note about the information it received, which is then attached to the person's landmark in the memory bank. Importantly, we request the VLM to use cardinal directions in the global map frame instead of relative directions such as ``left'' or ``right'' so that the note can be understood later without needing the agent's pose from the time of recording (Fig. \ref{fig:convo}).

\paragraph{Directional Sign (Sign Reading).} The agent approaches the sign and reads it via a call to the VLM. The sign text is grouped based on arrow direction (binned into cardinal directions), and the directions are transformed into the global map frame for recording (see Fig.~\ref{fig:overview} for example).

Individual prompts and further details for each type of VLM call can be found in Appendix~\ref{sup:prompts}.

\section{Experiments}
\label{sec:exp}

We evaluate \method{} across a suite of real and simulated navigation tasks. We seek to answer the following questions:
1). {\it Can our VLM leverage higher-order skills to avoid wrong searches?},
2). {\it How does \method{} perform in unseen real-world navigation tasks?},
3). {\it How do sign-reading and human interactions impact navigation efficiency?},
4). {\it How does map visualization input influence the VLM’s spatial understanding?}
We answer these questions through a variety of qualitative and quantitative analyses of the system's performance in comparison to relevant baselines.

\paragraph{Task description.} Our evaluation tasks are designed to mimic a realistic indoor delivery scenario. The agent is placed in a large unknown building and is tasked with finding a target room specified by a room number. The episode is considered successful if the target room label has been read by a VLM call, within a 15-minute time limit.

\paragraph{Real-world environment.} In the real-world, we consider two complex university campus multi-purpose buildings, each over 80m in length. For most of the rooms, including the target room, there is a room label next to each door to the room. Each building contains signs and people scattered throughout. The human responses are open-ended, but are typically along the lines of ``it's the second door down the hallway behind you.'' We evaluate navigation performance over 12 total trials in buildings A and B, each with different start and goal locations.

\begin{figure}[!ht]
    \centering
    \includegraphics[width=1\linewidth]{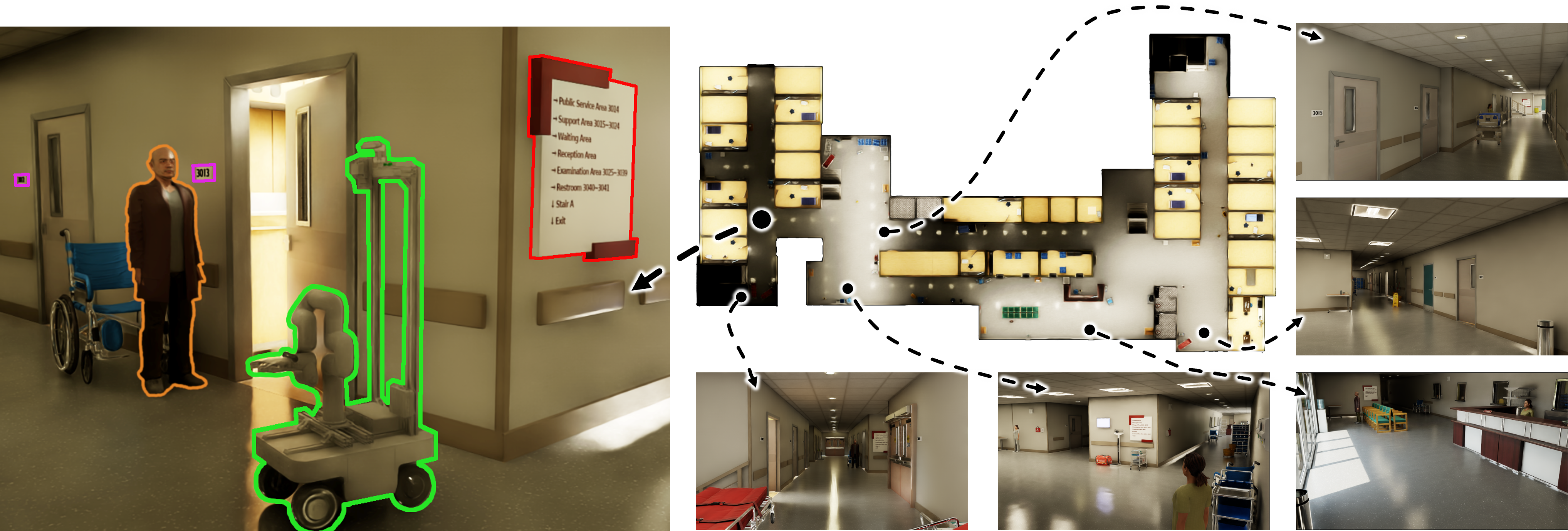}
   \caption{\textbf{Hospital Environment Visualization.} Existing open‐world navigation benchmarks do not support large‐scale building navigation tasks with human interaction. To fill this gap, we introduce an Isaac Sim‐based interactive navigation benchmark in a photorealistic hospital with over 30 rooms (offices, operation, examination, and patient rooms). The environment features realistic objects and layouts, informative signs, traversable rooms, and NPCs for human–robot interaction. We also provide a queryable website with an online staff directory.}  %
    \label{fig:sim_env}
    \vspace{-1em}
\end{figure}

\paragraph{Simulation environment.} We construct an environment in Isaac Sim simulation to enable reproducible evaluation (Fig.~\ref{fig:sim_env}). To the best of our knowledge, there is no existing simulation environment suitable for evaluating higher-order navigation skills in realistic man-made scenes. We use existing assets for an empty hospital and add room labels for each door, directional signs, and virtual humans who can provide directions via hard-coded conversational responses. We evaluate navigation performance over 14 trials, each with different start and goal locations. We plan to release all the code and assets needed for evaluation in this environment to accelerate future research on practical indoor navigation. More details are provided in Appendix~\ref{sec:sim_details}.

\paragraph{Baselines.}
To the best of our knowledge, there is no existing method available for navigating to specific rooms within buildings, as the task inherently requires integrating text reading capabilities into the navigation pipeline. Thus, we design baselines which can be thought of as ablations of our method. To determine the impact of higher-order navigation skills on navigation efficiency, we create a baseline in which signs and people are not processed into the landmark memory bank (No Signs/Humans Feedback, Fig. \ref{fig:ablation}). Thus, the VLM has no option to read signs or ask people for more information---it only sees map frontiers and doors and decides which to visit. We also experiment with removing the map image input to the VLM (No Landmark Map Input, Fig. \ref{fig:ablation}). In this case, the VLM only receives scene information via JSON text format. 

\paragraph{Metrics.}
We measure success rate, average episode duration, and average distance traveled. A success is counted when the robot reaches the goal and recognizes that it has completed the task after reading the room number. For failed episodes (due to collisions or timeout after 15 minutes), we assign a maximum duration of 900s and maximum distance traveled of 100m as a penalty.

\begin{figure}[!ht]
    \centering
    \includegraphics[width=1\linewidth,trim={0px 0px 0px 0px},clip]{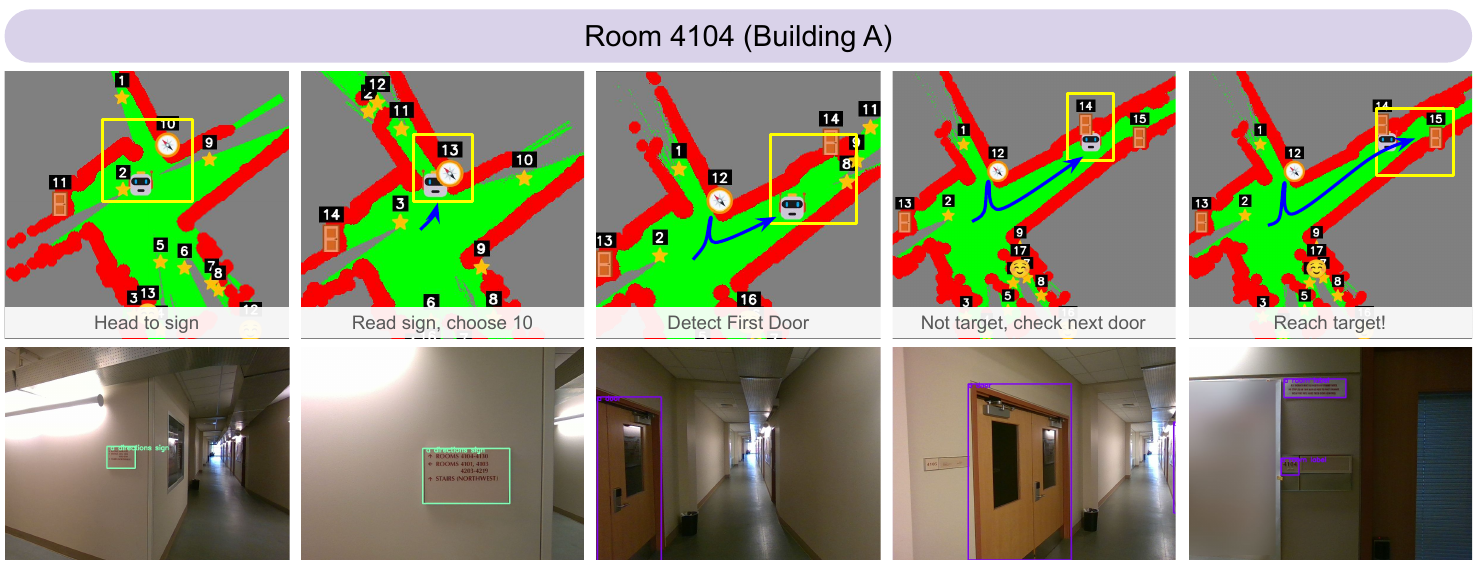} %
    \includegraphics[width=1\linewidth]{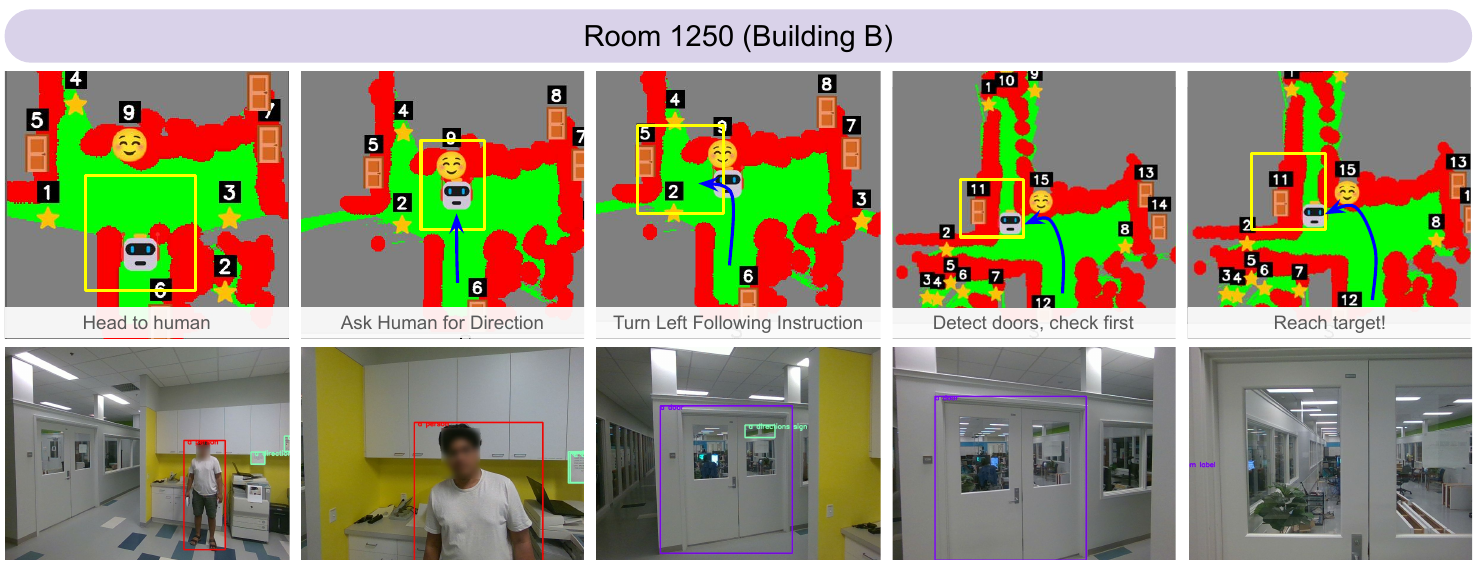}
    \caption{\textbf{Qualitative Results}: We present full step-by-step episode visualizations of our framework in two different real-world buildings. Thanks to its ability to reason over many sources of information, \method{} can accurately and efficiently navigate to the specified room number. Blue lines indicate the approximate traveled trajectories.}
    \label{fig:qualitative}
    \vspace{8pt}
\end{figure}

\subsection{Qualitative Results}
\label{sec:exp_qual}
We provide step-by-step episode visualizations of \method{}'s behavior in the real world in Fig.~\ref{fig:qualitative}. Note that in each example, there are landmarks in many different directions that the agent can choose from. Choosing to explore in a direction that does not lead to the goal may result in wasting time by exploring very long hallways. We observe that our VLM agent is able to successfully read signs, interpret their directions with respect to the provided map, and use the information to pick frontiers that directly lead to the goal. Similarly, the agent can ask people for directions, record the received information in its memory bank, and use it effectively in subsequent high-level planning steps.

\begin{figure}[!ht]
    \centering
    \includegraphics[width=0.9\linewidth]{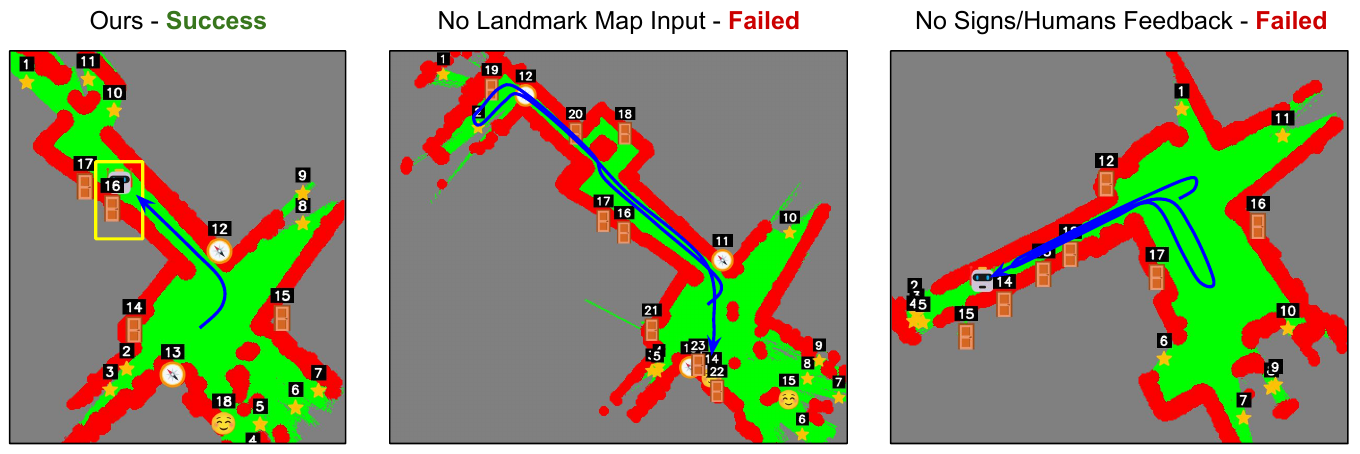}
    \caption{\textbf{Qualitative comparison with baselines.} We compare our method with ablative baselines to validate our visual prompting design and the importance of sign reading and communicating with humans. The visual map prompting enhances the spatial reasoning capabilities of the VLM, while the sign reading and communication gathers important information for efficient navigation.}
    \label{fig:ablation}
\end{figure}

We compare our method with the aforementioned baselines qualitatively in Fig.~\ref{fig:ablation}. Removing the map image input significantly hinders the VLM's spatial reasoning capabilities, making it more likely to misunderstand which doors are close to the agent and are worth visiting. This confirms that modern VLMs are able to interpret top-down map images and use them for planning. On the other hand, removing the ability to read signs and ask people for directions makes the agent more likely to go in a completely wrong direction, causing failure due to timeout. 

\subsection{Quantitative Results}
\label{sec:exp_quant}

We report quantitative results in both real-world and simulation environments (Tables~\ref{tab:success_rates_real_by_building} and~\ref{tab:success_rates_sim}). The results reveal several key insights. First, higher-order navigation skills—reading signs and asking people for directions—are critical: without these skills, the agent succeeds in only 8.3\% of real-world trials and 42.86\% of simulation trials. Second, omitting our unified spatial memory map as image prompts significantly degrades performance, underscoring its importance for decision making. In real-world experiments, baselines without skills or map prompting frequently exhaust their time budgets, illustrating that missing high-level cues leads to timeout failures. By combining sign-reading, direction-asking, and spatial-map integration, our approach outperforms both the “No Signs/People” and “No Map Image” baselines by over 40 percentage points—achieving 58.3\% in real-world trials and 57.14\% in simulation.

\begin{table}[!htb]
    \centering
    \caption{Quantitative Results for Navigation in Real-World Environments (Academic Complexes) }
    \label{tab:success_rates_real_by_building}
    \renewcommand{\arraystretch}{1.2}  %
    \begin{tabular}{lccccccc}
    \toprule
    & \multicolumn{2}{c}{Success Rate (\%)} 
    & \multicolumn{2}{c}{Avg Duration (s)} 
    & \multicolumn{2}{c}{Distance Traveled (m)} \\
    \cline{2-3} \cline{4-5} \cline{6-7}
    Environment      & Build A & Build B & Build A & Build B & Build A & Build B  & Avg. (\%)\\
    \hline
    No Signs/People  & 10         & 0         & 817.00       & 900         & 90.48         & 100     & 8.3   \\
    No Map Image     & 20         & 0         & 679.72         & 900         & 73.52         & 100     & 16.6    \\
    \textbf{Ours}    & \textbf{50}         & \textbf{100}         & \textbf{572.35}           & \textbf{232.63}        & \textbf{60.28}         & \textbf{12.61}    & \textbf{58.3}    \\
    \bottomrule
    \end{tabular}
\end{table}

\begin{table}[!htb]
    \centering
    \caption{Quantitative Results for Navigation in Simulation Environments (Large Hospital)}
    \label{tab:success_rates_sim}
        \renewcommand{\arraystretch}{1.2}  %
    \begin{tabular}{lccc}
    \toprule
    Environment & Success Rate (\%) & Duration (s) & Distance Traveled (m)  \\
    \hline
    No Signs/People   & 42.86 & 710.76 & 75.56  \\
    No Map Image      & 14.29 & 860.72 & 123.95 \\
    \textbf{Ours}     & \textbf{57.14} & \textbf{608.99} & \textbf{72.53} \\
    \bottomrule
    \end{tabular}
\end{table}

\subsection{Failure Analysis}
\label{sec:failure}

We provide an in-depth failure analysis of our method for both simulation and real-world scenarios in Table~\ref{tab:failure_causes}. Since failures in long-horizon navigation typically stem from multiple factors per episode, we include an analysis of the top 3 contributing causes for each failure. For failure cases with less than 3 contributing factors, all the factors are counted.

\begin{table}[h!]
    \centering
    \caption{Frequency of causes identified in top-3 failure reasons per episode.}
    \label{tab:failure_causes}
    \renewcommand{\arraystretch}{1.2}  %
        \begin{tabular}{lcc}
            \midrule
            \textbf{Failure Cause} & \textbf{Real Percentage}  & \textbf{Sim. Percentage} \\
            \midrule
            Incorrect Detection            & \textbf{33.33}  & 21.43 \\ 
            Detection Missed              & 26.67  & 21.43 \\
            Reasoning Failure              & 20.00  & 7.14 \\
            Incorrect Human Info           & 13.33  & 0.00  \\
            SLAM Failure                   & 6.67   & 14.29 \\
            Planner/Controller Failure                & 0.00   & \textbf{28.57} \\               
            \midrule
        \end{tabular}
\end{table}

In the real world, the detection module causes the most failures. Posters are often detected as directional signs or room labels, causing the robot to waste time reading them. In simulation, low-level navigation issues from the Nav2 module are most prevalent, especially where narrow hallways and obstacles make it difficult to plan paths, slowing the robot. In summary, improving low-level perception and planning performance are still important directions for future work.

\label{sec:exp_ablation}

\section{Conclusion}
\label{sec:conc}

We presented \method{}, a novel method for robot navigation that incorporates human-like navigation skills, such as sign reading and asking for directions, in an agentic VLM framework. \method{} abstracts low-level perceptual inputs into a memory bank of landmarks and uses a VLM to perform higher-order reasoning on these landmarks and plan high-level actions. We conduct experiments to validate the capabilities of the agent and show that higher-order navigation skills are important for efficient navigation in large buildings.

\clearpage

\section{Limitations}
\method{} successfully exhibits human-like navigation behaviors based on higher-order reasoning. However, since the system relies on an object detector to produce landmarks, the overall performance is bottlenecked by the detection performance. In addition, since the objects are limited to a predefined set of categories and the VLM only observes landmarks, the high-level planning does not maximally use the information contained in the camera observations. In the future, as detection capabilities become better integrated into VLMs themselves, the specialized detector could be replaced by a more powerful VLM-based detection stream. Lastly, because the VLM is restricted to exploring frontiers, it is not able to choose closer waypoints which may be sufficient for exploring an area while taking less time to reach. This could be mitigated by incorporating more sophisticated re-planning logic.

\acknowledgments{This project is supported by the Intel AI SRS gift, Amazon-Illinois AICE grant, Meta Research Grant, IBM IIDAI Grant, and NSF Awards \#2331878, \#2340254, \#2312102, \#2414227, and \#2404385. We greatly appreciate the NCSA for providing computing resources.}

\bibliography{main}  %

\newpage

\appendix

\section{Additional Experiments}
We provide additional comparisons to other ablative baselines to validate the usefulness of VLM reasoning and the JSON landmark memory bank (Table~\ref{tab:success_rates_ablation}). We tested on the same 14 simulation experiments as Table~\ref{tab:success_rates_sim}. For ``No VLM," we tested two heuristics: \textit{closest} and \textit{random} unvisited landmark selection. For ``No JSON," we removed the JSON landmark memory bank, leaving the landmark map as the only input to the VLM with scene information. We remark that since the JSON serves as the memory keeping track of (un)visited landmarks, this baseline keeps choosing visited landmarks, leading to time-out. 

\begin{table}[!htb]
    \centering
    \caption{Additional Ablations for Navigation in Simulation Environments (Large Hospital)}
    \label{tab:success_rates_ablation}
        \renewcommand{\arraystretch}{1.2}  %
    \begin{tabular}{lccc}
    \toprule
    Environment & Success Rate  \\
    \hline
    No VLM (closest)   & 35.71   \\
    No VLM (random)     & 28.57  \\
    No JSON & 0 \\
    \textbf{Ours}     & \textbf{57.14} \\
    \bottomrule
    \end{tabular}
\end{table}

\section{Simulation Environment Details}
\label{sec:sim_details}

Our simulation experiments were conducted using Isaac Sim with customized robot and environment assets. The robot is controlled via ROS 2 with the same topics/interfaces as the real-world robot to ensure easier sim-to-real transfer. 

\paragraph{Simulated Robot Setup}
Our simulation robot asset starts with a URDF exported from Fusion 360 CAD. The core sensors and actuators, including base control, pan-tilt, and sensor data (RGB-D camera, 3D LiDAR, and 2D LiDAR), are then simulated using Isaac Sim action graphs. Coordinate frames/topics are also aligned with the real-world robot system.

\paragraph{Simulated Environment Setup}
We modify one of the provided Isaac Sim hospital environments to include room numbers, signs, and 11 human NPCs for our task.  

\begin{figure}[!ht]
    \centering
    \includegraphics[width=1\linewidth]{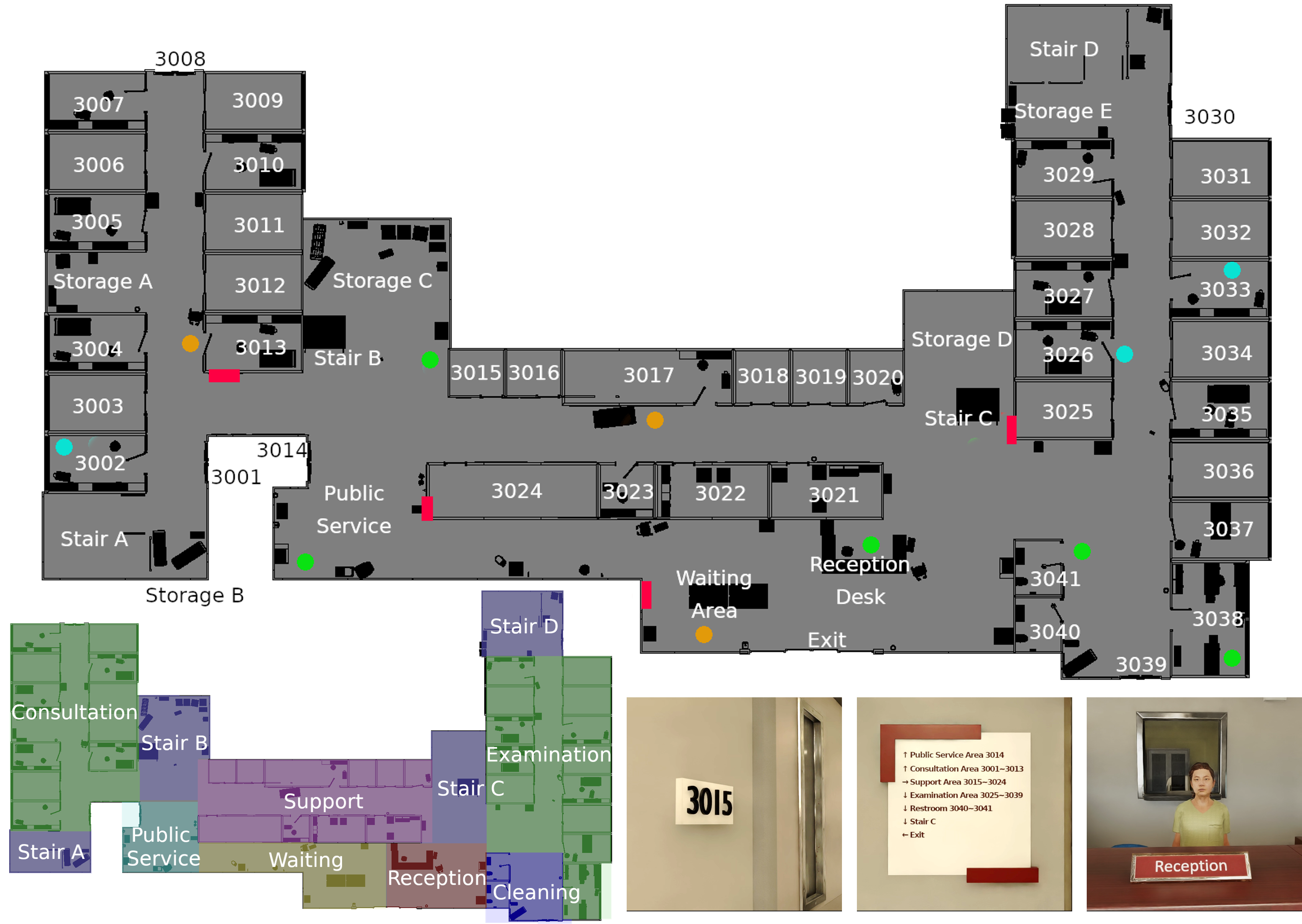} %
   \caption{\textbf{Door Numbering, Signs and NPC Placement in Hospital Environment.} Blue dots represent doctors, orange are patients, and green are nurses. Red rectangles represent signs.}  %
    \label{fig:sim_top_down}
\end{figure}

We divided the hospital into several functional zones: consultation, support, examination, reception, waiting, stairs, and public service areas. Each room within these zones was assigned a unique room number ranging from 3001 to 3041. Four prominent directional signs were placed at key locations to indicate the relative directions of room number ranges in each zone. The NPCs are arranged realistically in the scene as shown in Fig.~\ref{fig:sim_top_down}.

\paragraph{NPC Information}
To make the robot’s interaction with the environment more realistic, we classified NPCs into three types---doctors, nurses, and patients---each with its own `knowledge base.' \textbf{Doctors} know the exact locations of all doctors and approximate orientations of each functional zone. They can also answer medical questions and schedule consultations. \textbf{Nurses} know the exact locations of all doctors and all functional zones. They can also help with patient registration, check appointment times, and provide detailed directions to specific rooms. \textbf{Patients} know the approximate directions to the public service and waiting areas. Furthermore, NPCs provide hard-coded relative directions when asked (e.g., ``The room is on your left"). Each NPC has a different set of rooms that they know the directions to. If the NPC does not have information regarding the user's inquiry, they will reply with some version of ``Sorry, I don't know the answer to that question."

\section{Real-World Environment Details}
\label{sup:real_exp_setup}
We conducted 10 experiments in building A and 2 in building B for a total of 12 real-world experiments. In each of those experiments, one to two human actors were stationed at a fixed location to provide open-ended relative directions if asked. Non-actor humans will occasionally pass by.

\section{\method{} Implementation Details}
\label{sup:sysm_details}
 
\subsection{Robot Hardware Setup}
\label{sup:hardware}

\begin{figure}[h!]
    \centering
    \includegraphics[width=1\linewidth,trim={150px 110px 150px 110px},clip]{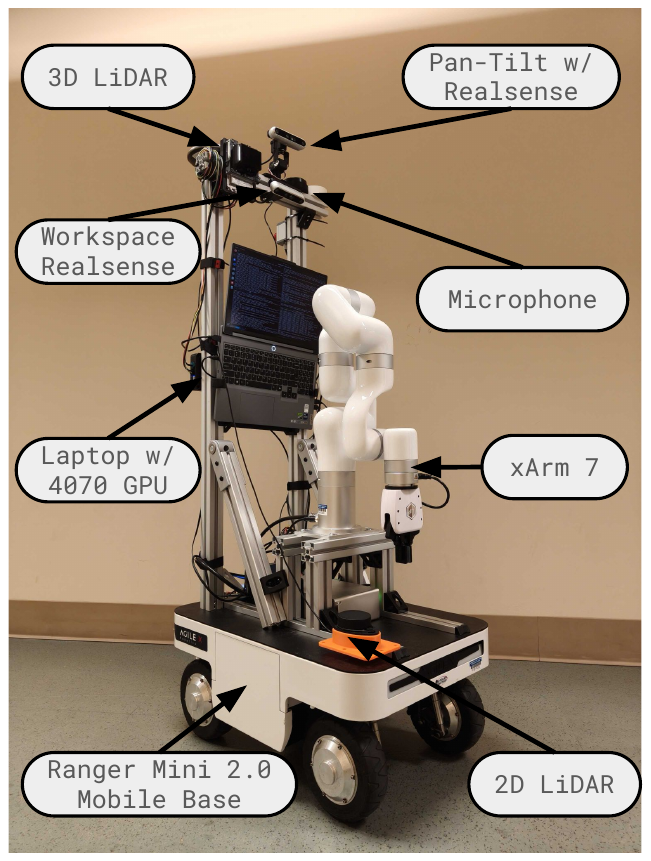}
   \caption{\textbf{Hardware System Overview.}}  %
    \label{fig:hardware}
    \vspace{-1em}
\end{figure}

Our robot is custom-built and consists of a mobile base, an arm, a computer, and various sensors. Components are attached to the base using aluminum t-slots and 3D-printed mounts. See Fig.~\ref{fig:hardware}.

We use the AgileX Ranger Mini 2.0, a mobile platform with 4-wheel steering and an onboard power supply that allows for omnidirectional motion in flat indoor and outdoor spaces. The 7-DoF UFACTORY xArm7 robot arm (currently unused) is mounted at a height ideal for manipulation on tables/door handles in the real world. For obstacle avoidance and mapping, there is a Slamtech 2D LiDAR mounted near the base and a Hesai FT120 solid-state 3D LiDAR mounted on top. Two Realsense D455's are mounted on top for perception, one on a pan-tilt mechanism and the other angled downward toward the robot workspace.
Only the pan-tilt camera is used in this work, and it is only used for object detection and text reading, not mapping. The height of the camera is roughly aligned with most indoor signs, and the pan-tilt capability allows for viewing the surroundings quickly without turning the entire robot. 
For interacting with humans, we use a Respeaker omnidirectional microphone, mounted near the top for conversation with humans. An Ardusimple simpleRTK3B GPS is also mounted but is unused in this work.

Our onboard compute comes from a Lenovo Legion 5i laptop with an NVIDIA 4070 GPU. Using an x86 laptop rather than an embedded ARM device like a Jetson allows for easier software setup and debugging. The entire system is integrated using ROS2 Humble and Docker, with all sensors/robot interfaces connected to the laptop. The laptop is connected to the internet in order to query GPT 4.1 via its API.

Overall, the system costs \(\sim \)\$35k. We plan on open-sourcing the entire CAD and hardware setup documentation in the future as well.

\subsection{Localization and Mapping}
\label{sec:lowlevel}
We first describe our system's low-level processing stream, which is responsible for producing a top-down map of the environment, localizing the agent with respect to the map, detecting certain objects, and path planning. We perform 2D simultaneous localization and mapping (SLAM) using SLAM Toolbox~\cite{slam_toolbox} which merges the 3D lidar scan into the 2D scan and performs optimization to produce a top-down occupancy map of areas the agent has explored. Concurrently, we perform object detection on images from the RealSense camera using an open-vocabulary detector without retraining. We use NanoOWL, an optimized implementation of OWL-ViT~\cite{minderer2022simple}, and query with three text labels:  \texttt{door}, \texttt{person}, and \texttt{directions sign}. We use a variety of heuristics to filter the detections and extract the 3D position and approach orientation for each detection (see supplementary for details). For path planning, we use NavFn, a wavefront Dijkstra planner from Nav2 ~\cite{macenski2020marathon2}, and execute the paths using an MPPI~\cite{williams2016aggressive} controller.

\subsection{Object Detection Details}
\label{sup:objdet}
The detection runs on a 15 FPS stream of $1280 \times 720$ RGB-D images from the RealSense camera. In the real-world, additional filtering is performed based on bounding box corners unprojected to 3D. For doors, the NanoOWL confidence threshold is $0.3$, and we only keep detections with box widths in the range $(0.5,2.5)$ and heights in the range $(0.5,3)$. %
For room labels, the confidence threshold is $0.04$, and the width range is $(0,0.4)$ and the height range is $(0,0.15)$. For directional signs, the confidence threshold is $0.03$, and the width range is $(0.35,0.5)$ and the height range is $(0.2,0.5)$. For people, the confidence threshold is $0.3$ and we do not do any other filtering.  The unprojected box center is taken as the 3D position of each object, and the approach orientation is acquired by taking the cross product of vectors from the center to unprojected points 20 pixels to the left and below. For each incoming frame, the filtered detections are greedily matched with existing objects based on center distance and aggregated if the distance is within 1m. Only the objects with 3 or more detections within the last 20 frames are added to the memory bank.   

\subsection{Frontier Extraction Details}
Following VLFM~\cite{yokoyama2024vlfm}, we compute frontiers as the midpoints of boundary segments separating explored and unexplored regions. First, a contour of the explored region is extracted using OpenCV's findContours function. The contour is broken into contiguous obstacle-free segments using a dilated obstacle map, and the midpoints are computed by traversing the path along each segment and accumulating the distance.

\subsection{Human Interaction Details}
We specify three types of human interaction: 1). getting instructions, 2). asking for directions, and 3). asking for directory information (e.g. what room Prof. Shenlong's office is in).

The first case happens at startup if a task is not yet specified. The robot will ask ``Hi, I am a delivery robot! How can I help you?" and wait for a response from the human. This response will then be passed into a VLM call to retrieve the goal in a specific format for parsing.

The second and third cases occur when the VLM planner chooses a human landmark. When asking for directions, the robot will say ``"Hi, I am a delivery robot! Do you know where \{goal\} is?". Here, directions are assumed to be given relative to the robot. After receiving the response, a VLM will be called to write a note to a future self on how to reach the goal using cardinal directions specified by the landmark map. This will be added to the JSON memory bank under the human landmark. When human responses are unclear (e.g. no response or a response that is not relevant), the corresponding field will mention that there is no information gained. Our system does not reason about which human has more information.

For getting directory info, the robot will ask ``Hi, I am a delivery robot! Do you know which room \# \{goal\} is in?" The response will be processed by a VLM and the navigation goal will be updated accordingly.

All of these interactions require human detection to initiate. Our current pipeline does not explicitly handle dynamic actors. If the VLM-selected human walks away, it will receive no response and thus gain no info. Robustly handling dynamic actors is left as an extension for future work.

\subsection{VLM Prompts}
\label{sup:prompts}
We provide the exact prompts for the various VLM calls we make in our framework: 1). Choosing the next landmark to visit given a top-down map image and JSON memory bank, 2). Reading the room number of a door given a real-world image, 3). Reading a directional sign given a real-world image, 4). Deciding which type of human interaction should occur, and 5). Recording a note given the speech-to-text transcription of a human interaction. All of them use the same system prompt.
\paragraph{System Prompt} 
\begin{quote}
    You are ChatGPT, a large language model trained by OpenAI.
    Follow the user's instructions carefully.
    Respond concisely, informatively, and helpfully.
    If you're unsure about an answer, say so.
    You have strong reasoning capabilities.
\end{quote}
\paragraph{1). Choose Next Landmark Prompt}
\begin{quote}
    You are a robot (\includegraphics[height=1em]{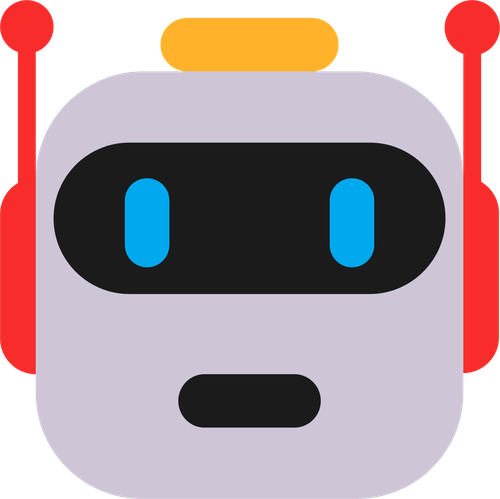}) trying to find \{target\}.
        
    The map depicts the environment you've explored so far with doors (\includegraphics[height=1em]{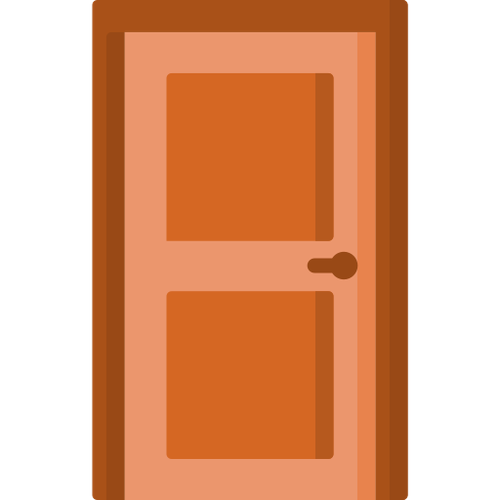}), signs (\includegraphics[height=1em]{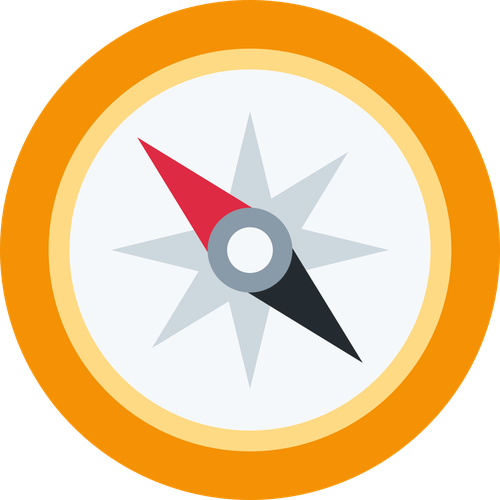}), people (\includegraphics[height=1em]{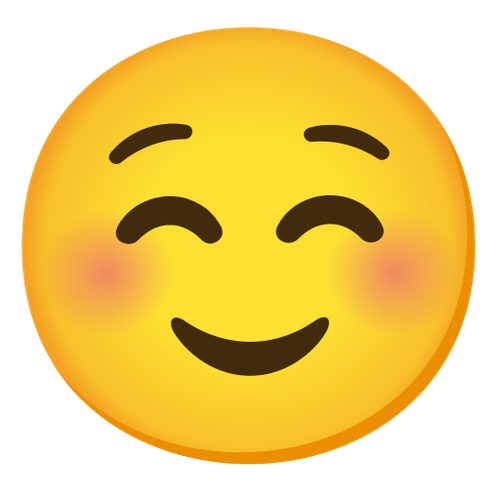}), and frontiers (\includegraphics[height=1em]{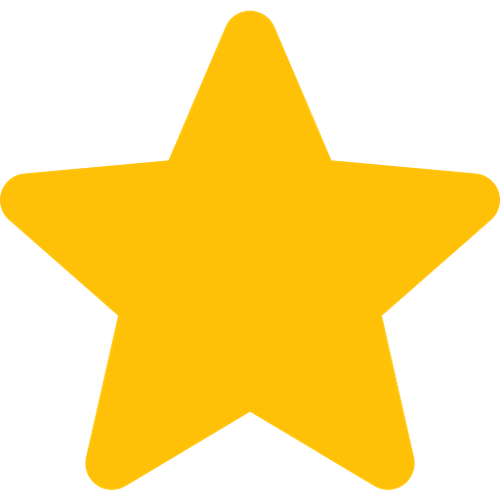}). The map is North up, and the borders are labeled with cardinal directions N, S, E, W.
            
    The JSON provided below maps each landmark index on the map to its name and, if applicable, directions (from signs) and info (from people).
    \\
    
    Choose the next landmark to go to based on the provided map and JSON.
    
    Reason over how to most **efficiently** find the target, like going to directions signs to narrow down regions, following ascending/descending door number patterns, visiting nearest doors to see their room number, asking a person nearby for directions, or exploring frontiers to find more landmarks.
    \\
    
    Visited doors are named in the format "Visited\_a door\_X", where X is the room number. Use these room numbers to identify door numbering patterns.
    
    Follow patterns efficiently. For example, in ascending patterns, if you are at door 100 and looking for door 110, skip doors 101, 102, etc.
    \\
    
    Keep in mind that all information from signs, people, etc. are relative to the position of that particular landmark, and is probably only applicable in a local region of that landmark.
    \\
    
    If an unvisited person is nearby and you want to ask for more information on where to go, choose the person. 
    Information provided by a visited person will be listed in 'info' using cardinal directions. Use this information.
    
    Example:
    
    \quad Input: target='Room 3339', info="I should go North to find the door"
    
    \quad Output: You will pick a landmark to the north of the image.
    \\
    
    If directions are available in the json, follow those directions. This will take precedence over looking for things nearby. Make sure your range is correct.
    
    Example: 
    
    \quad Input: target='Room 3339', directions={{'North':['Room 3326-3340'], 'North-East':[], 'East':[], 'South-East':[], 'South':['Room 3101-3307'], 'South-West':[], 'West':[], 'North-West':[]}}
    
    \quad Output: You will pick a landmark to the north of the image because 3339 is in the range 3326-3340. 
    
    **Do not choose landmarks that are already visited (i.e. names that look like "Visited\_obj\_id"**.
    \\\\
    First, think carefully step by step about where the target room might be and decide which landmark to visit next.  Then, print out your reasoning followed by the chosen landmark index in brackets.
    
    Example: I am looking for Room 110. The directions from the sign say room 100-120 are south. Landmark 3 is an unvisited door to the south, so it might be the target; Chosen landmark: [3]
    \\
    JSON: 
    
    \{vlm\_keypt\_dict\}
\end{quote}

\paragraph{2). Read Door Label Prompt}
\begin{quote}
    What is the door number in the image? Return only the number and confidence score (to 2 decimal places), separated by a semicolon. 
    
    If you can't see a sign or you can only read half of the sign, return -1; confidence score. 
    
    If you can see the sign but you can't read the numbers, return -2; confidence score. 
    
    Your confidence score should be between 0 and 1. Be more conservative with your estimates.
\end{quote}

\paragraph{3). Read Directional Sign Prompt}
\begin{quote}
    Break down the directions sign you see in the image.
    
    Return your answer in the form \{'left': [content], 'right': [content], 'forward': [content], 'backwards': [content]\}. Return only the dict (no comments or formatting).
\end{quote}

\paragraph{4). Choose Human Interaction Type Prompt}
\begin{quote}
    You are a delivery robot the current goal: \{goal\}.
    
    Given the information you have learned: \{learned\_info\},
    what do you want to ask the person in front of you?
    
    Pick the response that makes the most sense from the following 3 choices.
    
    Return the number and only the number.
    \\
    1. "How can I help you?" (This is used to get a goal, only use if the current goal is None)
    
    2. "Do you know where \{goal\} is?" (This is used to ask for directions)
    
    3. "Do you know which room \# \{goal\} is in?" (This is used to get directory information)
\end{quote}

\paragraph{5). Record Conversation Information Prompt}
\begin{quote}
    You are a robot (\includegraphics[height=1em]{media/robot_icon.png}) trying to find \{goal\}. You are facing \{robot\_facing\}. Given the map and the directions by Person, write a short note for your future self to refer to later on how to reach your goal.
    
    Return only the output in the format "Note: note to self". Do not include quotations.
    
    The image is already aligned with relative directions, so left means left on the image. In your note to self, use the cardinal directions in the map.
    
    Also ignore the numbers above each landmark as they will be updated.
    
    Conversation:
    \{conversation\_history\}
\end{quote}

\section{Additional Visualizations}
\label{sup:additional_viz}
\paragraph{Simulation Example}
We present a step-by-step qualitative result for our simulation environment in Fig. \ref{fig:sim_qual}, with more examples in the supplemental video. 
\begin{figure}[!ht]
    \centering
    \includegraphics[width=1\linewidth]{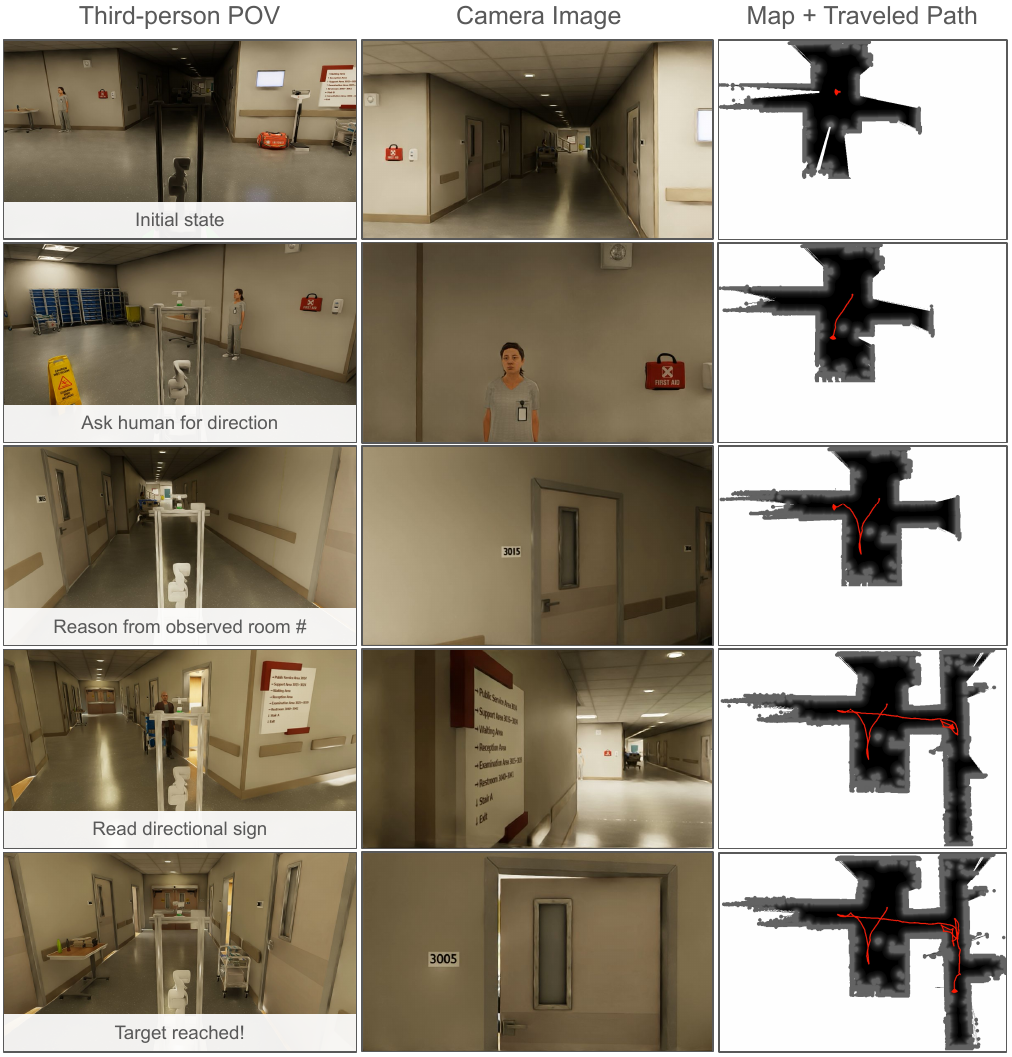}
    \caption{\textbf{Qualitative Simulation Results}: We present full step-by-step episode visualizations of our framework in simulation with exact path traveled highlighted in red.}
    \label{fig:sim_qual}
    \vspace{-8pt}
\end{figure}

\begin{figure}[!ht]
    \centering
    \includegraphics[width=1\linewidth]{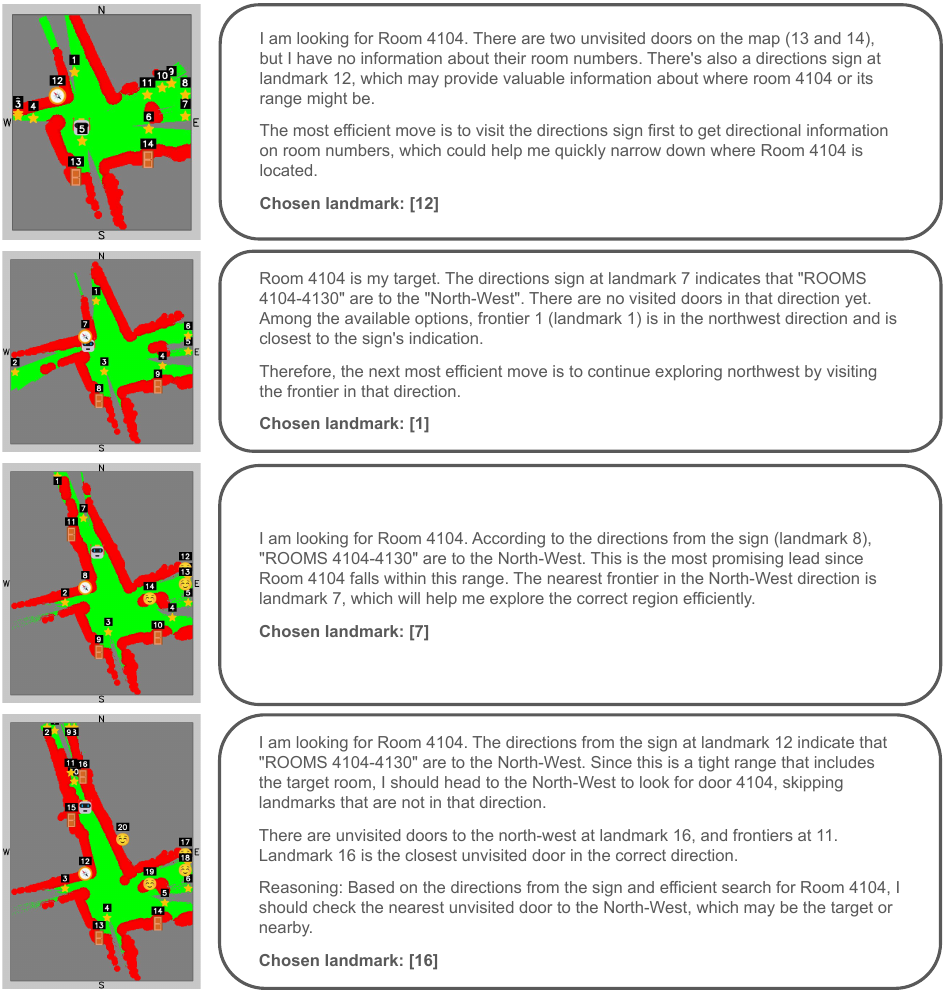}
    \caption{\textbf{Real-world VLM reasoning}: We present a step-by-step example of VLM's reasoning and decisions to navigate to room 4104. \method{} exhibits spatial reasoning capabilities given the direction guidance from direction signs, as showcased in the third and fourth rows.}
    \label{fig:vlm_reason}
    \vspace{-8pt}
\end{figure}

\paragraph{Step-by-step VLM reasoning example}
We further showcase \method{}'s reasoning capabilities in a step-by-step example in Fig. \ref{fig:vlm_reason}. Given the landmark map (on the left), the VLM is prompted to choose a waypoint for the robot to follow. The text boxes (on the right) represent the VLM's reasoning and decisions. Please refer to the video for more examples. 

\section{Extension to Multi-Floor Scenarios}
\label{sup:multifloor}
We show \method{} can also work with multiple floors through the elevator demo shown in the supplementary video at the 2:00 timestamp. We choose a goal room on the fourth floor and initialize the robot with the information that it is currently on the third floor and can take elevators. We add elevators and elevator button panels to our detection module as well, and design a simple behavior primitive to take the elevator. When the VLM chooses to go to the elevator, we detect the button panel and use our onboard robot arm to push the up button. After the button is pressed, the robot moves to face the door and uses the center depth values from the pan-tilt camera to sense when the elevator door has opened based on a threshold. The robot then moves forward and asks a human (assuming someone can help push the interior buttons) to push the button to go to the fourth floor. Using the pan-tilt camera facing backwards, the robot once again waits till the elevator door has opened before backing out. The SLAM and landmark maps get reset and the robot continues to search for the door as usual on the new floor.

\end{document}